%% file: arxive.tex
\documentclass[runningheads]{llncs}

\usepackage{eccv}

\usepackage{eccvabbrv}

\usepackage{graphicx}
\usepackage{booktabs}

\usepackage[misc]{ifsym}
\usepackage{tikz}
\usepackage{comment}
\usepackage{amsmath,amssymb} 
\usepackage{color}
\usepackage{multirow}
\usepackage[accsupp]{axessibility}  
\usepackage{float}
\usepackage{enumerate}
\usepackage{multicol}
\usepackage{makecell}

\usepackage[pagebackref,breaklinks,colorlinks,citecolor=eccvblue]{hyperref}

\usepackage{orcidlink}
\input{gqwritepackage}

\begin{document}

\title{CLII: Visual-Text Inpainting via Cross-Modal Predictive Interaction} 


\author{
Liang Zhao\inst{1}
\and
Qing Guo\inst{2}
\and
Xiaoguang Li\inst{1}
\and
Song Wang\inst{1}}

\authorrunning{Liang Zhao, Qing Guo, Xiaoguang Li, Song Wang}

\institute{University of South Carolina, USA \and IHPC and CFAR, Agency for Science, Technology and Research (A*STAR), Singapore \\
\email{\{lz4,xl22\}@email.sc.edu, tsingqguo@ieee.org, songwang@cec.sc.edu}}

\maketitle

\begin{abstract}
  Image inpainting aims to fill missing pixels in damaged images and has achieved significant progress with cut-edging learning techniques.
    Nevertheless, state-of-the-art inpainting methods are mainly designed for nature images and cannot correctly recover text within scene text images, and training existing models on the scene text images cannot fix the issues. 
    In this work, we identify the \text{visual-text inpainting} task to achieve high-quality scene text image restoration and text completion: Given a scene text image with unknown missing regions and the corresponding text with unknown missing characters, we aim to complete the missing information in both images and text by leveraging their complementary information. 
    Intuitively, the input text, even if damaged, contains language priors of the contents within the images and can guide the image inpainting. Meanwhile, the scene text image includes the appearance cues of the characters that could benefit text recovery.
    To this end, we design the cross-modal predictive interaction (CLII) model containing two branches, \ie, ImgBranch and TxtBranch, for scene text inpainting and text completion, respectively while leveraging their complementary effectively.
    Moreover, we propose to embed our model into the SOTA scene text spotting method and significantly enhance its robustness against missing pixels, which demonstrates the practicality of the newly developed task.
    To validate the effectiveness of our method, we construct three real datasets based on existing text-related datasets, containing 1838 images and covering three scenarios with curved, incidental, and styled texts, and conduct extensive experiments to show that our method outperforms baselines significantly.
\end{abstract}

\begin{figure*}[t]
  \centering
   \includegraphics[width=1.0\linewidth]{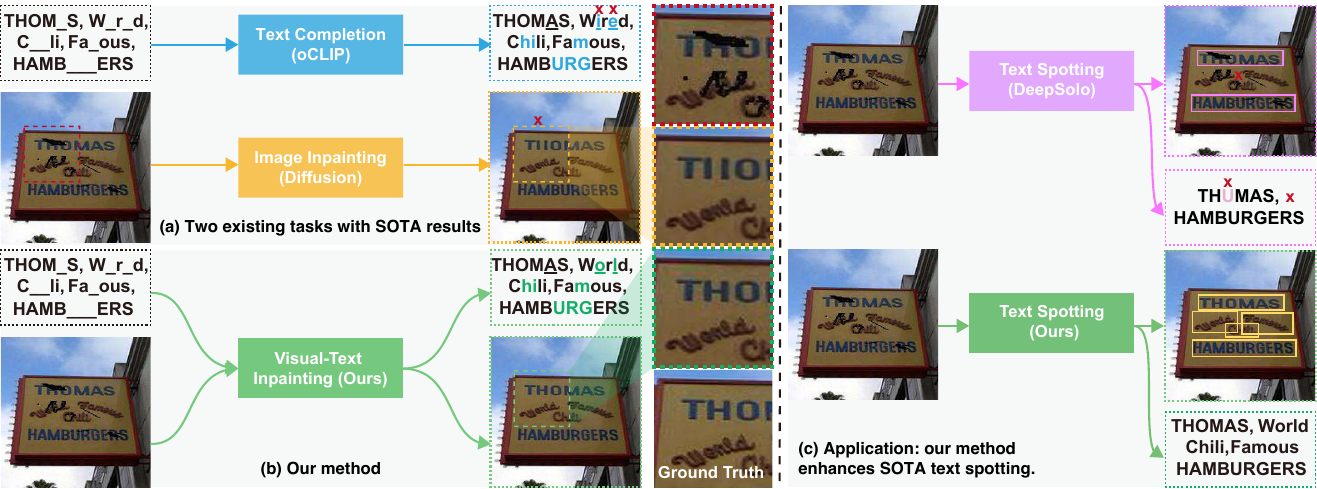}
   \caption{Visual-text inpainting vs. existing tasks. (a) Existing image inpainting and text completion methods, respectively. (b) The proposed visual-text inpainting. (c) Leveraging our method to enhance SOTA text spotting method \cite{ye2023deepsolo} We highlight the key distinctions within dashed boxes.}
   \label{fig:fig1}
\end{figure*}

\section{Introduction}
\label{sec:intro}

In urban streets, texts are widely printed on different materials, such as advertising boards, walls, shop signs, warning signs, etc., to achieve the purpose of publicity and informing.
Identifying the texts within captured images accurately could help understand the scene (\eg, text translation), in particular, it is useful for travelers visiting a city with an unknown language.
However, printed texts could be easily corrupted in the real world due to occlusion or material aging.
Hence, it is critical to achieve robust text recognition.
A potential solution for damaged text identification is first to detect the text in the damaged image and then complete the missing characters.
Nevertheless, this solution has two limitations: (1) Extracting text contents from a damaged image is difficult. As shown in \figref{fig:fig1} (c), the SOTA text spotting method \cite{ye2023deepsolo} cannot detect the corrupted text regions and fail to recognize most of the words. (2) The text completion may raise error predictions. As displayed in \figref{fig:fig1} (a), the SOTA text completion method mispredicts the word `world.'
Another potential solution is to leverage image inpainting to reconstruct the damaged image first and then extract the text.
Image inpainting is a fundamental computer vision problem that fills missing pixels in a natural image. State-of-the-art works \cite{li2022misf, li2020recurrent, guo2021jpgnet, yeh2017semantic, li2023leveraging, zhang2023superinpaint, zhang2023sair} leveraging deep generative models can realize high-quality restoration.
However, reconstructing text contents in the image denoted as scene text inpainting remains unexplored and has inherent challenges that existing image inpainting methods can hardly solve: 
Given a damaged scene text image, the scene-text inpainting aims to reconstruct the image backgrounds and the text contents, intricately linked to language priors.
State-of-the-art image inpainting methods can hardly get good results since the text could encompass an infinite array of character combinations featuring diverse fonts and styles.
For example, in \figref{fig:fig1} (a), the state-of-the-art diffusion-based inpainting method can fill the missing pixels to get a natural image but fails to recover text accurately: characters `H' becomes `l l'.


Instead of performing scene text inpainting directly, it is relatively easier to complete missing characters in a given text string by leveraging the language priors in language models. 
As shown in the \figref{fig:fig1} (a), given the same text, we use the state-of-the-art text completion method to fill in the missing characters, which can complete most of the characters but with some minor errors.
Intuitively, the two tasks, \ie, plain text completion and scene-text inpainting, are complementary due to the inherent connection between the two modalities.
Inspired by this phenomenon, we explore end-to-end modeling to complete the missing information in both modalities and identify a new task as \textit{visual-text inpainting} (See \figref{fig:fig1} (b)).

To achieve this goal, we propose a novel model (\ie, cross-modal predictive interaction (CLII)) to leverage the complementary information within two modalities effectively.
Specifically, CLII consists of two branches for scene text inpainting and text completion, which are denoted as ImgBranch and TxtBranch, respectively. 
Different from performing the two tasks independently, we propose to leverage the latent embedding from TxtBranch containing the language priors to enhance the embedding of ImgBranch and then decode it for reconstruction. 
Meanwhile, we adopt the latent embedding from ImgBranch to benefit the text decoding in TxtBranch.
With CLII, we can not only complete characters in the text modal but also generate high-quality images with well-reconstructed damaged characters (See \figref{fig:fig1} (b)). 
Moreover, we equip our method with the state-of-the-art scene text spotting method (\ie, \cite{ye2023deepsolo}) and enhance its robustness against damages significantly (See \figref{fig:fig1} (c)).
Our main contributions could be summarized as follows:
\begin{itemize}
    \item We identify the visual-text inpainting task aiming to complete the missing pixels and characters of a given scene-text image and the corresponding text.

    \item We propose a novel model, \ie, cross-model predictive filtering (CLII), to realize visual-text inpainting, which can complete the missing pixels and words effectively and simultaneously.

    \item We equip our CLII with the state-of-the-art scene-text spotting method and enhance its robustness significantly.

    \item We present an insightful discussion on the visual-text inpainting task and conduct extensive evaluations on three challenging datasets. Our method outperforms state-of-the-art methods by a large margin. 
    
\end{itemize}
%

\section{Related Work}
\label{sec:relatedwork}

{\bf{Deep Generative Image Inpainting}} Instead of finding useful patches for the restoration of damaged images, recent endeavors have adopted deep generative networks to acquire more nuanced information for inpainting. Iizuka \etal~\cite{iizuka2017globally} explore the local and global consistency of the damaged regions, 
while Yan \etal~\cite{yan2018shift} and Yu \etal~\cite{yu2018generative} establish correlations among long-range regions. Moreover, Liu \etal~\cite{liu2018image} employ spatial deformations within the damaged regions, and Nazeri \etal~\cite{nazeri2019edgeconnect} as well as Ren \etal~\cite{ren2019structureflow} leverage edge and contour information for inpainting. 
To leverage the Transformer, He \etal~\cite{he2022masked} mask random patches and reconstruct images from latent representation and mask tokens. Cao \etal~\cite{cao2022learning} extend ViT \cite{he2022masked} with informative priors to enhance the long-distance dependencies. Recently, Li \etal~\cite{li2022misf} propose multi-level siamese filtering to fill large missing areas with details. These methods collect more contextual information from images to improve inpainting details. These methodologies excel in gathering contextual information from images, but text content or damaged characters cannot be directly obtained from the scene contexts.

{\noindent \bf{Diffusion Inpainting.}} With the realm of diffusion models \cite{avrahami2023blended,rombach_high-resolution_2022,lugmayr2022repaint}, text-guided diffusion inpainting \cite{xie2023smartbrush,nichol2021glide,avrahami2022blended}, facilitates image editing in accordance with user-specified instructions. An example involves a prompt such as ``replace the chair with a stool'',  wherein the diffusion model eliminates the chair object and introduces a stool. Nevertheless, it is essential to emphasize the marked disparity between our task and the mentioned methods. Our task has no sentences for text inpainting. There are no corresponding objects in the scene aligned with the damaged characters, which cannot provide text content for inpainting. 
Recent diffusion inpainting methods \cite{wang2022zero,zhang2023towards,lugmayr2022repaint} utilize the diffusion process in reconstructing natural images. However, the content of images is hard to control, let alone the text content.


\noindent {\bf{General Text Recognition.}} Enlightened by Natural Language Processing (NLP), text recognition endeavors to classify each character within words based on visual features. The sequence information is processed by the Recurrent Neural Network (RNN) and is matched with vocabulary. Attention-based text recognition methods incorporate with the variations of RNN, such as canonical attention BLSTM \cite{shi2016robust,cheng2018aon,liu2018char,zhan2019esir}, bidirectional attention BLSTM \cite{shi2018aster,chen2020adaptive}, two-dimension attention BLSTM \cite{li2019show,yang2019symmetry,wang2019scene,baek2019wrong} and Transformer decoders \cite{zhu2019text,bartz2019kiss,fang2021read,atienza2021vision}. These sequence processing models leverage one or two-dimensional visual features to comprehend character relationships in a unidirectional or bidirectional manner, enhancing text recognition.  
In essence, text recognition harnesses complete visual information to manage character relationships and generate recognized texts, though the performance is much worse when confronted with damaged texts. In contrast, our approach diverges by utilizing damaged visual information to predict missing characters, relieving the recognition errors on damaged texts and also facilitating visual-text inpainting.

{\noindent \bf{Damaged Text Completion.}} Conventional language approaches within the NLP community, exemplified by models like CLIP \cite{radford2021learning}, BERT \cite{devlin2018bert}, and their variations, specialize in predicting missing words within sentences or absent sentences within a paragraph. These methodologies are designed to discern word relationships within the context of sentences. When incorporated with vision tasks, the words always correspond to the objects or behaviors inside images. However, the completion of missing characters or damaged texts remains inadequately studied. Additionally, character contents have no semantic corresponding counterparts in images that could serve as cues for text completion. The closest related work \cite{xue2022language} employs a character-aware text encoder and decoder to predict the masked characters of input texts. It aims to learn character priors to enhance text recognition performance. In contrast, our approach leverages damaged text images to predict the missing characters and also inpaint the missed characters with visual attributes, which retrieve priors for both character contents and visual designs through the proposed visual-language embedding method.

\begin{figure*}[t]
  \centering
   \includegraphics[width=0.9\linewidth, height=0.32\linewidth]{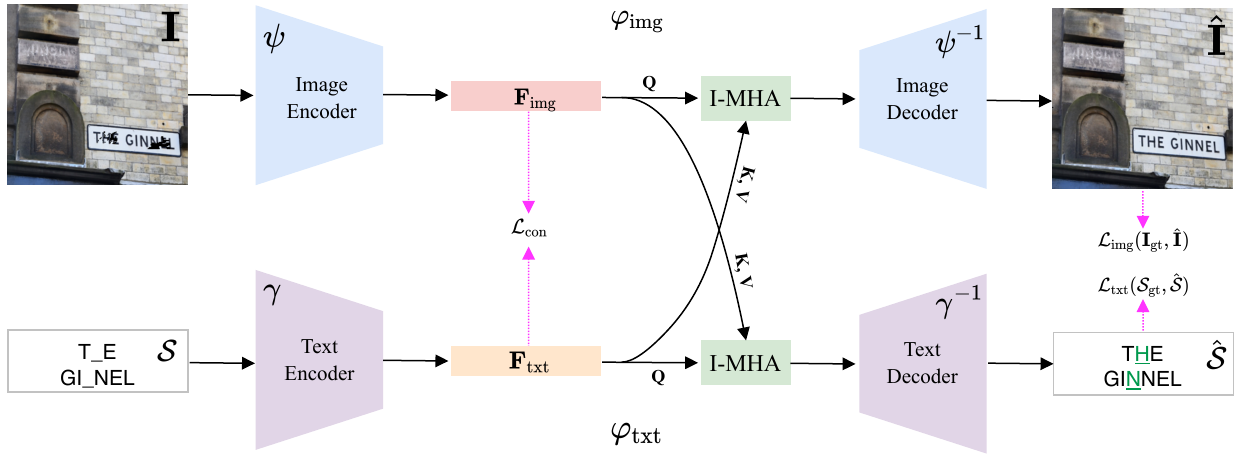}
   \caption{Architecture of the proposed method. The visual text and plain text are randomly masked, respectively. ``MHA'' denotes the Multi-Head Attention.}
   \label{fig:framework}
\end{figure*}

\section{Visual-Text Inpainting}
\label{sec:visual-text}

We first formulate the visual-text inpainting problem and detail our cross-modal predictive interaction (CLII) in \secref{subsec:CLII} with detailed implementations in \secref{subsec:train}. 

\subsection{Problem Formulation}
\label{subsec:problem}

We have a scene text image $\mathbf{I}$ and a text string $\mathcal{S}$, which contain the same text contents, but both are damaged with unknown missing regions in $\mathbf{I}$ and unknown missing characters in $\mathcal{S}$ (See \figref{fig:framework}).
Visual-text inpainting aims to recover both modalities by filling in missing pixels and characters.
We can represent the task as follows for easy understanding,
\begin{align}
    (\hat{\mathbf{I}},\hat{\mathcal{S}}) = \phi(\mathbf{I},\mathcal{S}), \text{subject to,}~ \text{Str}(\hat{\mathbf{I}})=\hat{\mathcal{S}},
\end{align}
where $\hat{\mathbf{I}}$ and $\hat{\mathcal{S}}$ are the restored image and text string, and the constraint $ \text{Str}(\hat{\mathbf{I}})=\hat{\mathcal{S}}$ means that the restored image should contain the same words or characters with the completed text string $\hat{\mathcal{S}}$.   
This task's main challenge is leveraging the complementary information within the two modalities effectively.
For example, in \figref{fig:fig1} (a), the character `H' in the input image is masked, while the damaged string contains the completed `THOM' that can be used to guide the inpainting of `H' in the image.  
Similarly, the completed characters in the image could be used to supervise the completion of the missing characters in the damaged string.

\subsection{Cross-Modal Predictive Interaction}
\label{subsec:CLII}

\subsubsection{Framework overview.}
CLII contains two branches, \ie, ImgBranch and TxtBranch, which are represented as $\varphi_\text{img}()$ and $\varphi_\text{txt}()$, respectively. In particular, ImgBranch aims to extract the embeddings of the input image and reconstruct the image based on the embeddings of the input text string, and we can formulate the whole process as
\begin{align} \label{eq:imgbranch}
   (\hat{\mathbf{I}}, \mathbf{F}_\text{img}) = \varphi_\text{img}(\mathbf{I}, \mathbf{F}_\text{txt}),
\end{align}
where $\mathbf{F}_\text{txt}$ is the embeddings from the text string and the embeddings $\mathbf{F}_\text{img}$ is used to guide the text completion.
That is, we can set the TxtBranch as
\begin{align} \label{eq:txtbranch}
    (\hat{\mathcal{S}}, \mathbf{F}_\text{txt}) = \varphi_\text{txt}(\mathcal{S}, \mathbf{F}_\text{img}).
\end{align}
In the following, we detail the two branches.

\subsubsection{ImgBranch $\varphi_\text{img}(\cdot)$.}

We extract the image's embedding by
\begin{align} \label{eq:imgbranch_encoder}
   \mathbf{F}_\text{img} = \psi(\mathbf{I}),
\end{align}
where $\psi(\cdot)$ is the encoder network. To fuse the embeddings of text (\ie, $\mathbf{F}_\text{txt}$), we propose to build an interaction transformer that consists of $N$ interactive multi-head attention layers (I-MHAs). 
For the $i$th I-MHA layer, we have the following formulation
\begin{align}\label{eq:imgbranch-imha}
    \mathbf{Q}_{i+1} & = \text{FFN}_i(\text{I-MHA}_i(\mathbf{Q}_i, \mathbf{K}_i, \mathbf{V}_i)) \nonumber \\                  & = \text{FFN}_i(\text{softmax}(\frac{\mathbf{Q}_i \mathbf{K}^\top_i}{\sqrt{d_k}})\mathbf{V}_i),
\end{align}
where the key and value are set as text modality's embedding $\mathbf{K}_i=\mathbf{V}_i = \mathbf{F}_\text{txt}$ and $\mathbf{Q}_{i}=\mathbf{F}_\text{img}$. 
Then, we can get $\mathbf{F}_{\text{img}\leftarrow \text{txt}} = \mathbf{Q}_{N}$ and decode it for image reconstruction
\begin{align} \label{eq:imgbranch-decoder}
    \hat{\mathbf{I}} = \psi^{-1}(\mathbf{F}_{\text{img}\leftarrow \text{txt}}),
\end{align}
where $\psi^{-1}(\cdot)$ is the decoder. In summary, ImgBranch is formulated as
\begin{align} \label{eq:imgbranch-sum}
    \hat{\mathbf{I}} = \varphi_\text{img}(\mathbf{I}, \mathbf{F}_\text{txt}) =  
 \psi^{-1}(\text{I-Transfer}(\psi(\mathbf{I}), \mathbf{F}_\text{txt})),
\end{align}
where $\text{I-Transfer}(\cdot)$ denotes the interaction transformer defined in \reqref{eq:imgbranch-imha}.

\subsubsection{TxtBranch $\varphi_\text{txt}(\cdot)$.}
We adopt a character-level text encoder to extract the embedding of characters. Assuming that the input text string consists of $T$ substrings, we can represent it as $\mathcal{S}=[\mathcal{S}_1, \mathcal{S}_2, ..., \mathcal{S}_T]$. 
The $t \text{th}$ text string $\mathcal{S}_t$ consists of maximum $K$ characters and is denoted as
$\mathcal{S}_t = [\mathbf{c}^t_1, \mathbf{c}^t_2, ..., \mathbf{c}^t_{K}]$. 
We embed each character in a text string $\mathcal{S}_t$ as a character vector $\mathbf{F}^t_k$, 
\begin{align} \label{eq:txtbranch-en1}
    \mathbf{F}^t_k = \mathbf{W} \cdot \mathbf{c}^t_k + \mathbf{P}_k,  t \in [1, T], k \in [1, K],
\end{align}
where $\mathbf{W}$ is the character embedding matrix and $\mathbf{P}_k$ is the learnable positional encoding \cite{xue2022language}. The damaged characters are masked. Then, we can get text embeddings $\mathbf{F}_\text{txt}$,
\begin{align} \label{eq:txtbranch-en2}
    \mathbf{F}_\text{txt} = \gamma(\mathcal{S}) =  \text{concat} ([\text{MHA}_k(\mathbf{F}^t_k)]), t \in [1, T], j \in [1, {K}],
\end{align}
where character relationships are encoded by $\text{MHA}$ transformer and all text instances are concatenated by $\text{concat}(\cdot)$ as text embeddings. Similar to the \reqref{eq:imgbranch-imha}, to fuse the embeddings of image (\ie, $\mathbf{F}_\text{img}$), we adopt an interaction transformer that consists of $N$ interactive multi-head attention layers (I-MHAs). 
For the $i$th I-MHA layer, we have
\begin{align}\label{eq:txtbranch-imha}
    \mathbf{Q}_{i+1} & = \text{FFN}_i(\text{I-MHA}_i(\mathbf{Q}_i, \mathbf{K}_i, \mathbf{V}_i)) \nonumber \\                  & = \text{FFN}_i(\text{softmax}(\frac{\mathbf{Q}_i \mathbf{K}^\top_i}{\sqrt{d_k}})\mathbf{V}_i),
\end{align}
where $\mathbf{K}_i=\mathbf{V}_i = \mathbf{F}_\text{img}$ and $\mathbf{Q}_{i}=\mathbf{F}_\text{txt}$. 
Then, we can get $\mathbf{F}_{\text{txt}\leftarrow \text{img}} = \mathbf{Q}_{N}$ and decode it for text completion
\begin{align} \label{eq:txtbranch-decoder}
    \hat{\mathcal{S}} = \gamma^{-1}(\mathbf{F}_{\text{txt}\leftarrow \text{img}}),
\end{align}
where  $\gamma^{-1}(\cdot)$ is the text decoder. In summary, we have the TxtBranch as
\begin{align} \label{eq:txtbranch-sum}
    \hat{\mathcal{S}} = \varphi_\text{txt}(\mathcal{S}, \mathbf{F}_\text{img})
    = \gamma^{-1}({\text{I-Transfer}}(\gamma(\mathcal{S}), \mathbf{F}_\text{img})),
\end{align}
where $\text{I-Transfer}(\cdot)$ denotes the interaction transformer defined in \reqref{eq:txtbranch-imha}, and $\gamma(\cdot)$ denotes the encoding process of text defined in \reqref{eq:txtbranch-en2}.

\subsection{Network Training and Implementation Details}
\label{subsec:train}

The training coordinates the optimization of both branches of ImgBranch and TxtBranch. We detail loss functions and datasets as follows.

\subsubsection{Loss functions.}
For the text completion, we adopt cross-entropy recognition loss during training. 
\begin{align}\label{eq:losstxt}
    \mathcal{L}_{\text{txt}} (\mathcal{S}_\text{gt}, \hat{\mathcal{S}}) = \text{H} (\mathcal{S}_\text{gt}, \hat{\mathcal{S}}),
\end{align}
where $\mathcal{S}_\text{gt}$ are string ground truths, $\hat{\mathcal{S}}$ are the predicted strings, and \text{H} is the cross-entropy function. For the image inpainting, we use the $\mathcal{L}_1$ loss to measure the normalized distance of reconstructed images with the ground truth. 
\begin{align}\label{eq:lossimg}
    \mathcal L_\text{img}(\mathbf{I}_\text{gt}, \hat{\mathbf{I}}) = \| \mathbf{I}_\text{gt} - \hat{\mathbf{I}} \| _1,
\end{align}
where $\mathbf{I}_\text{gt}$ is the ground truth and $\hat{\mathbf{I}}$ is the reconstructed image. To align the visual information with text instances, we adopt the contrastive loss \cite{xue2022language}. 
For a batch during the training process, we have $B$ text strings and images, respectively, denoted as $(\mathbb{S} = \{\mathcal{S}_i\}_{i=1}^B, \mathbb{I}=\{\mathbf{I}_j\}_{j=1}^B)$, which can form $B^2$ combinations and $(\mathcal{S}_i,\mathbf{I}_j|i=j)$ have the same text contents.
Then, given an image $\hat{\mathbf{I}}$ and a text string $\hat{\mathcal{S}}$, we can calculate the similarity of $\hat{\mathbf{I}}$ to the $i$th text $\mathcal{S}_i$ and the similarity of $\hat{\mathcal{S}}$ to the $j$th image $\mathcal{I}_j$ by
\begin{align}\label{eq:lossconstra}
    \mathbf{p}^\text{i2t}_i(\hat{\mathbf{I}}) = \frac{\text{exp}(\hat{\mathbf{I}}, \mathcal{S}_i)}{\sum ^B _{b=1} \text{exp}(\hat{\mathbf{I}}, \mathcal{S}_i)},~~\text{and}~~
    \mathbf{p}^\text{t2i}_j(\hat{\mathcal{S}}) = \frac{\text{exp}(\hat{\mathcal{S}}, \mathbf{I}_j)}{\sum ^B _{b=1} \text{exp}(\hat{\mathcal{S}}, \mathbf{I}_j)}.
\end{align}
After that, we can calculate the similarity between $\hat{\mathbf{I}}$ and each text string in $\mathbb{S}$ by
$\{\mathbf{p}^\text{i2t}_i(\hat{\mathbf{I}})|i\in [1,\ldots,B]\}$ that forms a vector denoted as $\mathbf{p}^\text{i2t}\in \mathds{R}^{B\times 1}$.
We can also count the similarity between $\hat{\mathcal{S}}$ and each image in $\mathbb{I}$ by
$\{\mathbf{p}^\text{t2i}_b(\hat{\mathcal{S}})|j\in [1,\ldots,B]\}$ and get a vector $\mathbf{p}^\text{t2i}\in \mathds{R}^{B\times 1}$.
We set the one-hot similarity ground truth $\mathbf{p}^\text{i2t}(\mathbf{I}_\text{gt})$ and $\mathbf{p}^\text{t2i}(\mathcal{S}_\text{gt})$ as $0$ for negative pairs and $1$ for positive pairs. 
We have a contrastive loss: 
%
\begin{align}    
    \mathcal L_\text{con}(\hat{\mathbf{I}}, \hat{\mathcal{S}}) =  \text{H} (\mathbf{p}^\text{i2t}(\mathbf{I}_\text{gt}), \mathbf{p}^\text{i2t}(\hat{\mathbf{I}})) + \text{H} (\mathbf{p}^\text{t2i}(\mathcal{S}_\text{gt}), \mathbf{p}^\text{t2i}(\hat{\mathcal{S}})).  
\end{align}
Finally, we have the overall optimization objective is:
\begin{align}
    \mathcal L = \lambda _1 \mathcal{L}_{\text{txt}} (\mathcal{S}_\text{gt}, \hat{\mathcal{S}})
    + \lambda _2 \mathcal{L}_\text{img}(\mathbf{I}_\text{gt}, \hat{\mathbf{I}})
    + \lambda _3 \mathcal{L}_\text{con}(\hat{\mathbf{I}}, \hat{\mathcal{S}}) .
\end{align}

\subsubsection{Training dataset.} We train our model on synthetic data SynthText \cite{Gupta16} and evaluate our method on other datasets to avoid any overfitting risk. SynthText dataset contains synthetic word instances of 800K images.

\subsubsection{Implementation details.}
The input images are resized to 256 $\times$ 256 and masked on texts with randomly chosen masks. The optimizer is Adam \cite{kingma2014adam} with decoupled weight decay regularization. The initial learning rate is
$1e^{-4}$ and uses a cosine schedule \cite{loshchilov2016sgdr} decay. Each text instance's maximum length is 25 \cite{xue2022language}. The depth of the Transformer for Image-Text interaction is 6. The number of encoder and decoder layers is 3 and 2, respectively. The model is trained end-to-end on 2 Tesla-V100 GPUs with a batch size 16. The model predicts 94 classes, including 52 upper and lower-case characters, 10 numbers, and 32 other punctuation symbols. The loss weights $\lambda _{rec}$, $\lambda _{c}$, and $\lambda _{l1}$ are set to 1.0, 1.0, and 1.0, respectively. 

\section{Leveraging CLII for Robust Scene-Text Spotting}
\label{sec:r-sts}

In this section, we detail how to use our method to enhance the robustness of the state-of-the-art scene-text spotting method, \ie, DeepSolo \cite{ye2023deepsolo}. To make it clear, we first introduce the DeepSolo, and then we explain how to embed our method. 

Given an image $\mathbf{I}$, DeepSolo predicts the text positions and recognitions the characters within it. First, it uses a transformer-based encoder $\tau(\cdot)$ to extract the embeddings of $\mathbf{I}$. Then, we map the embeddings to the position queries (\ie, $\mathbf{P}_q$) and text content queries (\ie, $\mathbf{C}_q$), which form the composite queries $\mathbf{Q}_q$. After that, we decode it via $\tau^{-1}$ to generate the recognition. The whole process can be formulated as
\begin{gather} \label{eq:spotting}
   \mathbf{S} =\text{MLP} (\tau ^{-1} (\mathbf{Q}_q)), 
   \mathbf{Q}_q = \text{PE}(\tau (\mathbf{I}))+\mathbf{C}_q,
\end{gather}
where $\mathbf{S}$ is the recognized text string. For a clean image, DeepSolo can recognize texts accurately. 
However, if the input image $\mathbf{I}$ is damaged, DeepSolo can hardly recognize and locate the texts accurately (See \figref{fig:fig1} (c)). 

We propose using our method to enhance the robustness of DeepSolo against missing pixels. 
Specifically, given a damaged image $\mathbf{I}$ (See \figref{fig:fig1} (c)), we first employ the DeepSolo to get the initial text string $\mathcal{S}$ where some characters are missed. 
Then, we feed $\mathbf{I}$ and $\mathcal{S}$ input our method (\ie, \reqref{eq:imgbranch} and \reqref{eq:txtbranch}) and output the restored image $\hat{\mathbf{I}}$ and $\hat{\mathcal{S}}$. 
After that, we further input $\hat{\mathbf{I}}$ and $\hat{\mathcal{S}}$ to the DeepSolo and generate the final text localization and recognition results.
We display the results in \figref{fig:fig1} (c) and quantitative comparison in  \secref{subsec:deepsolo}.
Our method enhances the robustness of DeepSolo significantly.

\section{Experiments}

\subsection{Setups}
\label{subsec:setups}

\subsubsection{Datasets.}
\label{sssec:datasets}
We evaluate our method on three datasets, \ie, Total-Text \cite{ch2020total}, ICDAR2015 \cite{karatzas2015icdar}, and TextSeg \cite{xu2021rethinking}. Total-Text dataset includes arbitrarily shaped scene texts. ICDAR2015 is an incidental scene text benchmark with small texts. 
TextSeg is a pixel-level annotated dataset with various text styles and designs. The distinguishing attributes of the evaluation benchmarks can verify the robustness of our method. 

\subsubsection{Metrics.} 
Following the previous image inpainting methods \cite{li2022misf}, we use the peak signal-to-noise ratio (PSNR) and structural similarity index (SSIM) to assess the reconstruction quality of our method's ImgBranch. Additionally, we employ the Precision (P) metric to evaluate the TxtBranch's performance.

\subsubsection{Baselines.} 
%
%
To demonstrate the effectiveness of our method in both text completion and image inpainting tasks, we compare it with the SOTA text completion method \cite{xue2022language} and image inpainting methods MAE \cite{he2022masked}, MAE-FAR \cite{cao2022learning}, MISF \cite{li2022misf}, and DDNM \cite{wang2022zero} separately. 
Additionally, we form baseline comparisons by combining existing image inpainting and text completion methods, \ie, MAE+oCLIP, MAE-FAR+oCLIP, MISF+oCLIP, and MISF*+oCLIP, where '*' indicates that the method is trained on SynthText dataset from scratch. 
To showcase our method's capability in enhancing text spotting, we also compare it with the SOTA text spotting method DeepSolo \cite{ye2023deepsolo}.

\subsection{Comparison Results}

\subsubsection{Quantitative comparison.}
We compare our method with competitors across three widely used datasets: Total-Text, ICDAR2015, and TextSeg. The result is shown in tab \ref{tab:stacomparison}. When conducting the experiments, we set the image mask ratio to 20\% and the text mask ratio up to 20\%. 
From tab \ref{tab:stacomparison}, we can observe that: \ding{182} 
{\bf For the text completion task,} damaging the image can significantly impact the performance of text completion. Specifically, when randomly masking out 20\% of an image, the precision of oCLIP \cite{xue2022language} decreases by 19.04\%, 11.83\%, and 18.49\% on the Total-Text, ICDAR2015, and TextSeg datasets, respectively.
Comparing our method with oCLIP \cite{xue2022language}, our method demonstrates its superiority across all three datasets by a significant margin, specifically, by 36.92\% in the Total-Text dataset, 33.33\% in the ICDAR2015 dataset, and 26.25\% in the TextSeg dataset.
\ding{183}{\bf For the image inpainting task,} our method outperforms all competitors across three datasets. Specifically, when comparing it with the transformer-based image inpainting method MAE \cite{he2022masked}, our method increases the PSNR and SSIM by 88.82\% and 30.96\%, respectively, on the Total-Text dataset. When compared with newly MAE-FAR \cite{cao2022learning}, our method increases the PSNR and SSIM by 69.58\% and 29.06\%, respectively on the Total-Text dataset. When compared with another image inpainting method MISF \cite{li2022misf}, our method increases the PSNR and SSIM by 2.21\% and 0.05\%, respectively, on the ICDAR2015 dataset. 
When compared with the diffusion-based method DDNM \cite{wang2022zero}, our method increases the PSNR and SSIM by 6.01\% and 2.86\%, respectively, on the TextSeg dataset.
\ding{184} {\bf For the formed baselines,} we first reconstruct the degraded image using existing image inpainting methods, \ie, MAE \cite{he2022masked}, MAE-FAR \cite{cao2022learning}, MISF \cite{li2022misf}, and DDNM \cite{wang2022zero} and then use the completed image along with the masked text as input for oCLIP \cite{xue2022language}. Finally, we evaluate both the image reconstruction quality and text completion performance. Compared to inputting the degraded image directly, employing the reconstructed image as input significantly improves the text completion performance of oCLIP \cite{xue2022language}. Specifically, MISF+oCLIP outperforms oCLIP (masked) by 7.33\% in terms of precision. When comparing our method with the formed baselines, we observe significant performance improvements in both image inpainting and text completion. Specifically, our method outperforms MISF$^*$+oCLIP by 23.59\%, 3.47\%, and 0.06\% in precision, PSNR and SSIM, respectively, on the Total-Text dataset. They are 34.31\%, 1.80\% and 0.05\% on the ICDAR2015 dataset. They are 6.41\%, 4.79\% and 0.12\% on the TextSeg dataset.
\begin{table*}[t]
  \centering
  \caption{Comparison results on Total-Text, ICDAR2015 and TextSeg dataset. }
  \renewcommand\arraystretch{1.4} 
  \resizebox{1.0\linewidth}{!}{
  \begin{tabular}{l|l|ccc|ccc|ccc}
    \toprule
	\multirow{2}{*}{Category} &\multirow{2}{*}{Methods} & \multicolumn{3}{c|}{Total-Text} &\multicolumn{3}{c|}{ICDAR2015} &\multicolumn{3}{c}{TextSeg}\\ 
    \cline{3-11}
     &  &P &PSNR &SSIM &P &PSNR &SSIM &P &PSNR &SSIM \\
    \hline
	\multirow{2}{*}{\makecell{Text \\  Completion}} 
    & \scriptsize{oCLIP} \scriptsize{\cite{xue2022language}} (\scriptsize{original}) &61.82 &- &- &51.32 &- &-  &75.95 &-&-\\
    & \scriptsize{oCLIP} \scriptsize{\cite{xue2022language}} (\scriptsize{masked}) &50.05 &-&-&45.25 &-&-&61.91 &-&-\\
    \hline
    \multirow{3}{*}{\makecell{Image \\ Inpainting}} 
    &\scriptsize{MAE} \scriptsize{\cite{he2022masked}} &- &21.4133 &0.7597 &- &26.4899 &0.8231  &-&19.3489 &0.6892\\
    &\scriptsize{MAE-FAR} \scriptsize{\cite{cao2022learning}} &- &23.8425 &0.7709 &- &27.3173 &0.8241  &-&22.4899&0.6974\\
    &\scriptsize{MISF} \scriptsize{\cite{li2022misf}}  &- &38.7840 &0.9938 &- &49.9073 &0.9990  &-&36.5758&0.9915\\
    &\scriptsize{DDNM} \scriptsize{\cite{wang2022zero}}  &- &37.0490 &0.9920 &- &50.7131 &0.9986  &-&36.5683&0.9908\\
    \hline
    \multirow{3}{*}{Formed baselines}
    & \scriptsize{MAE+oCLIP} &47.49&21.4133&0.7597 &43.98&26.4899&0.8231 &58.51&19.3489&0.6892 \\
    & \scriptsize{MAE-FAR+oCLIP} &50.79&23.8425&0.7709 &44.11&27.3173&0.8241 &62.03&22.4899&0.6974 \\
    & \scriptsize{MISF+oCLIP} &53.72 &38.7840 &0.9938 &44.87&49.9073&0.9990 &69.24&36.5758&0.9915 \\
    & \scriptsize{MISF*+oCLIP}  &55.45  &39.0773 &0.9943 & 44.92  &50.1074 &0.9990   &73.45  &36.9045  &0.9924 \\
    \hline 

    & Ours &$\textcolor{red}{68.53}$ &$\textcolor{red}{40.4326}$ &$\textcolor{red}{0.9949}$ &$\textcolor{red}{60.33}$ &$\textcolor{red}{51.0110}$ &$\textcolor{red}{0.9995}$ &$\textcolor{red}{78.16}$ &$\textcolor{red}{38.7663}$ &$\textcolor{red}{0.9936}$ \\
    \bottomrule
  \end{tabular} }
  \vspace{-18pt}
  \label{tab:stacomparison}
\end{table*}

\subsubsection{Qualitative comparison.}
\label{subsec:visualcomp}
\begin{figure*}[h]
  \centering
   \includegraphics[width=1.0\linewidth]
   {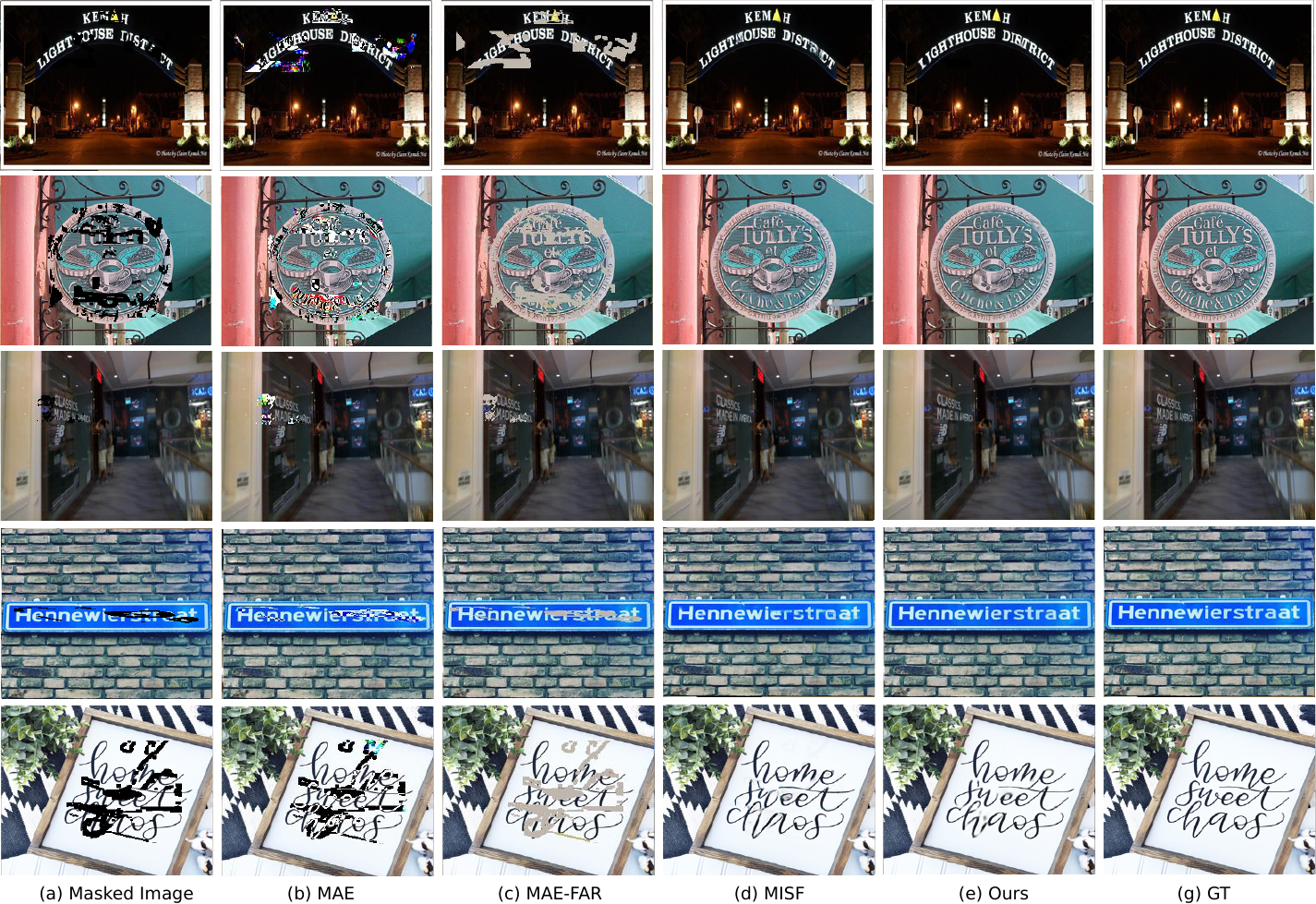}
   \caption{Qualitative results on Total-Text, ICDAR2015, and TextSeg. From left to right, the columns are masked images, MAE \cite{he2022masked}, MAE-FAR \cite{cao2022learning}, MISF\cite{li2022misf}, our results, and ground truth.}
   \vspace{-10pt}
   \label{tab:vis}
\end{figure*}

The visualization results of text inpainting are shown in Fig. \ref{tab:vis}. The first two rows are from Total-Text with arbitrarily shaped and curved texts; the third row is from the ICDAR2015 dataset with extremely unnoticeable texts; and the last two rows are from TextSeg of styled texts. 
Specifically, the MAE \cite{he2022masked} and MAE-FAR \cite{cao2022learning} struggle to discern the surrounding scene to recover the images, albeit without text content. MISF \cite{li2022misf} exhibits an ability to reconstruct some damaged texts but falls short in accurately restoring primary character strokes. Instead, our method excels in nearly eliminating most mask trails and reconstructing the texts with intricate character details. Specifically, in the last second row of example from TextSeg dataset, MAE and MAE-FAR could not figure out the inpainting in the masks and the trails are still obvious. The result of MISF could remove masks and fill with appropriate pixels. However, the character ``a'' is recovered as ``o''. Our method could recover the correct character.


\subsection{Discussion}
\label{subsec: discuss}

\subsubsection{Advantages of visual-text inpainting over existing tasks.}
\label{sssec:vti-vs-existing}
With two damaged modalities (\ie, text and image), there are three related tasks: text completion, image inpainting, and our task visual-text inpainting. 

\begin{wrapfigure}{r}{0.5\textwidth}
  \centering  
  \includegraphics[width=1.0\linewidth]
  {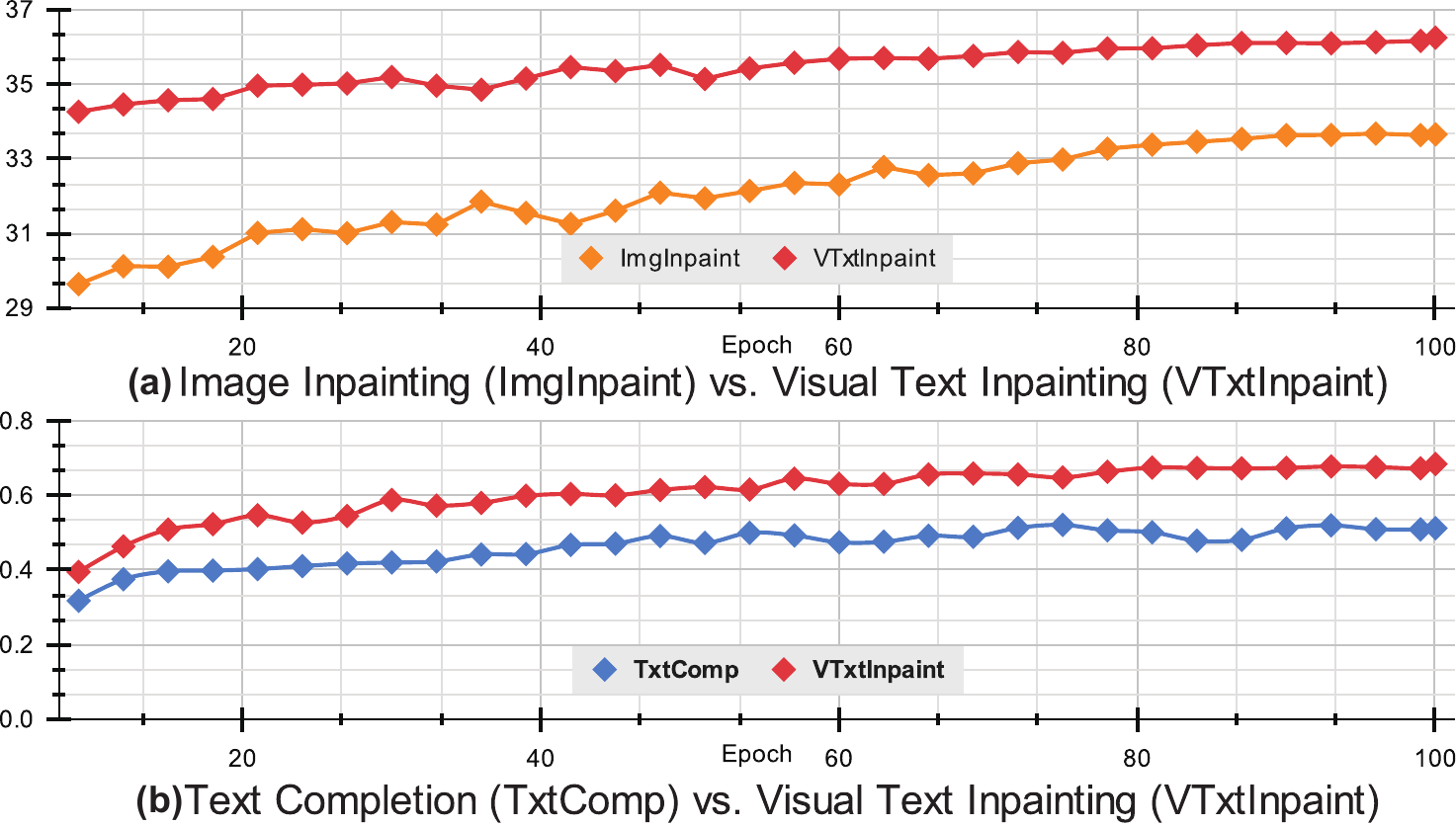}
   \caption{Performance comparison of three tasks.}
   \vspace{-19pt}
   \label{fig:taskcmp}
\end{wrapfigure}

To better understand the advantages of our task, we study the three tasks on the same dataset and evaluate their performance based on the corresponding metrics.
Specifically for the text completion and the image inpainting 
tasks, we adopt the state-of-the-art (SOTA) methods, \ie, oCLIP \cite{xue2022language} and MISF \cite{li2022misf}.
In terms of the visual-text inpainting, we adopt our model.
We train the three models on the SynthText dataset for a maximum of 100 epochs and test them on the Total-text dataset. 
We show the results in \figref{fig:taskcmp} and observe that our method for visual-text inpainting outperforms the image inpainting task and text completion task with a large margin consistently.

\subsubsection{Effectiveness of image cues for text completion.}
%
\begin{wraptable}{r}{0.5\textwidth}
  \centering
  \vspace{-22pt}
  \small
  \caption{Validating image cues for text completion with precision metric.}
  \vspace{5pt}
  \begin{tabular}{l|c|c|c}
    \toprule
    Input & oCLIP & oCLIP+  & CLII \\ 
    \hline
    Clean image   & 60.43  &63.60  & 65.67  \\
    Masked image  & 50.95 & 56.59 & 65.53 \\
    \bottomrule
  \end{tabular} 
  \vspace{-15pt}
  \label{tab:effectsofImg2Txtcmp}
\end{wraptable}
%
We aim to validate that our method can embed image information for text completion.
Specifically, we compare three methods including (1) oCLIP: plain text completion with oCLIP \cite{xue2022language} that takes a clean image and masked text as inputs for training and testing. 
(2) oCLIP+: modified text completion based on oCLIP. To validate our method, we build an enhanced oCLIP method by training oCLIP with masked image and text pairs.
(3) Our method CLII is also trained with masked image and text pairs.
We train and test the three models on the SynthText dataset. During testing, we consider two setups: clean and masked input images.
\begin{wrapfigure}{r}{0.5\textwidth}
  \centering
   \vspace{-12pt}
   \includegraphics[width=1.0\linewidth]
   {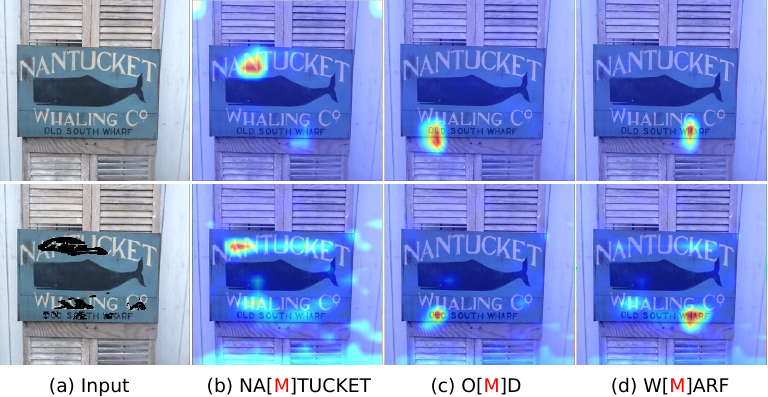}
   \caption{Comparison of attention maps. ``M'' in red color is the damaged character.}
   \vspace{-15pt}
   \label{fig:attn}
\end{wrapfigure}

We report the precision results in \tableref{tab:effectsofImg2Txtcmp} and show that: \ding{182} With extra image inputs, oCLIP+ and our method CLII get much higher precisions than the original oCLIP significantly, which demonstrates that the image information can benefit the text completion.
\ding{183} Comparing CLII with oCLIP+, CLII obtains much higher precisions under both clean and masked inputs. This demonstrates the effectiveness of our cross-modal interaction method.
\ding{184} With masked images, oCLIP and oCLIP+ get worse results than the models using clean images. In contrast, our method CLII gets similar results with clean and masked images, demonstrating our method's advantages.
To better understand the effectiveness, we calculate the attention maps of CLII via the method used in oCLIP under clean and masked images with different masked texts. As shown in \figref{fig:attn}, the attention maps have minor changes between clean and masked images under different masked texts, demonstrating that our interaction method is robust to missing pixels.

\begin{wraptable}{r}{0.5\textwidth}
  \centering
  \vspace{-20pt}
  \caption{Mutual influence between ImgBranch and TxtBranch.} 
  \label{tab:alldamage}
  \resizebox{1.0\linewidth}{!}{
  \begin{tabular}{l|c|ccc}
    \toprule
	\multirow{2}{*}{Image mask ratios} & \multirow{2}{*}{Text mask ratios} & \multicolumn{3}{c}{CLII} \\ 
    \cline{3-5}
    & &P &PSNR &SSIM \\
    \hline 
    \multirow{2}{*}{0\% $\sim$ 20\%}  &0\%$\sim$20\% &68.53 &40.4326 &0.9949  \\
      &20\%$\sim$40\% &50.74 &40.3060 &0.9943  \\
    \hline
    \multirow{2}{*}{20\% $\sim$ 40\%}  &0\%$\sim$20\% &64.32 &33.1509 &0.9798  \\
      &20\%$\sim$40\% &45.47 &32.0787 &0.9702  \\
      
    \hline
    \multirow{2}{*}{40\% $\sim$ 60\%}  &0\%$\sim$20\% &55.28 &28.2038 &0.9493 \\
     &20\%$\sim$40\% &39.42 &27.0978 &0.9383 \\
    \bottomrule
  \end{tabular}
  }
  \vspace{-20pt}
\end{wraptable}
\subsubsection{Mutual influence between image inpainting and text completion.} 
To explore the mutual influence between image inpainting and text completion, we conduct experiments with the following setup: (1) We mask the input image with three different mask ratios, namely, 0\%–20\%, 20\%–40\%, and 40\%–60\%. (2) For each image mask ratio, we randomly mask out input characters with two different mask ratios, namely, 0\%–20\% and 20\%–40\%. We report the experiment results in Tab. \ref{tab:alldamage} and find that:
\ding{182}{\bf The input image significantly affects the text completion performance.}
As the image mask ratio increases, there is a significant performance decrease observed in text completion. Damaged images could provide effective information for text completion. For each category of image damage, text completion encounters a larger fluctuation than text inpainting. For example, when image damages are serious up to 60\% and large parts of text are destroyed, the rest visual texts can only provide limited information for text completion. When the image damages are shallow around 20\% and large parts of texts are reserved, the visual texts can provide effective information in predicting the missing characters. 
For all categories, large image damages experience a serious degradation of text completion.
\ding{183}{\bf The input text significantly affects the image inpainting quality.} 
The greater the extent of damaged texts, the more pronounced the degradation in text inpainting. For example, for image damages are up to 20\%, PSNR decreases from 40.4326 to 40.3060 when text damages from around 20\% to around 40\%. For image damages are up to 40\%, PSNR decreases from 33.1509 to 32.0787 when text damages from around 20\% to around 40\%. For image damages are up to 60\%, PSNR decreases from 28.2038 to 27.0978 when text damages from around 20\% to around 40\%. It indicates that damaged texts can also provide important information for image inpainting, though texts are relatively small in whole images.

\subsubsection{Applications in scene text spotting.}
\label{subsec:deepsolo}

We further evaluate the reconstruction performance in the SOTA scene text spotting method, DeepSolo \cite{ye2023deepsolo}, as shown in Tab. \ref{tab:spotting}. 
``None'' denotes recognition without a vocabulary for both Total-Text and ICDAR2015 datasets. ``Full'' involves the vocabulary for Total-Text. ``S'', ``W'', and ``G'' correspond to vocabulary types of strong, weak, and generic, respectively, for ICDAR2015 dataset. Each vocabulary comprises varying numbers of vocabulary entries. With the help of vocabulary, wrong predictions of texts could be replaced with closest correct word in the vocabulary. This is an assistant technique in existing scene text spotting methods, while in real-world applications there might be less of vocabulary, especially for historical or specifically designed words. The metric is the F-score of text recognition in percentage values. 

From the results, we can see a notable degradation in text recognition when the images are damaged. The ``clean'' images have no damages and the recognition is 79.81\% and 70.37\% for Total-Text and ICDAR2015 datasets, respectively. When the images are damaged, which is common in practical situations, the performance decreases to 29.91\% and 8.42\%, respectively. However, upon correction by our method, the F-score for recognition exhibits a clear improvement, particularly for non-vocabulary results. Specifically, our text completion could improve the recognition from 29.91\% to 38.21\% on Total-Text dataset. For ICDAR2015 dataset, it is from 8.42\% to 11.06\%.  
Moreover, upon reconstructing the damaged images, which is the full application of our method, the scene text spotting results demonstrate approximately twofold improvements in the case of damaged images. Specifically, the recognition performance increases from 29.91\% to 54.86\% on Total-Text dataset. It increases from 8.42\% to 20.29\% on ICDAR2015 dataset. The vocabulary related results also show significant improvements. While some words might not be totally corrected, they can match with the vocabulary and get much higher recognition performance for further usage. The results demonstrate the effectiveness of our method in text related applications.  

\begin{table*}[t]
  \centering
  \caption{Scene text spotting comparison on Total-Text and ICDAR2015 datasets.  }
  \setlength{\tabcolsep}{4pt}
    \resizebox{1.0\linewidth}{!}{
  \begin{tabular}{l|cc|cccc}
    \toprule
	\multirow{2}{*}{Methods}  & \multicolumn{2}{c|}{Total-Text} &\multicolumn{4}{c}{ICDAR2015} \\ 
    \cline{2-7}
     &None &Full &None &S &W &G  \\
    \hline	
    DeepSolo \cite{ye2023deepsolo} (clean) &79.81 &86.97 &70.37 &86.81 &81.87 &76.69\\
    \hline
    DeepSolo \cite{ye2023deepsolo} (masked) &29.91 &45.21 & 8.42 &15.70 &13.32 &10.59 \\    
    DeepSolo$\mathbf{+}$Ours (masked)  &38.21 &45.42 &11.06 &15.71 &13.71 &11.93\\
    DeepSolo$\mathbf{+}$Ours (inpainting) &$\textcolor{red}{54.86}$ &$\textcolor{red}{67.83}$ &$\textcolor{red}{20.29}$ &$\textcolor{red}{33.33}$ &$\textcolor{red}{29.20}$ &$\textcolor{red}{24.77}$ \\
    \bottomrule
  \end{tabular} }
  \label{tab:spotting}
\end{table*}

\section{Conclusion}
\label{sec:conclusion}

We have introduced an interactive method for scene text inpainting and text completion, which can complete the missing pixels and characters at the same time.
To leverage the complementary information, we proposed the cross-modal predictive interaction containing two branches to handle the two modalities, respectively.
Moreover, we have embedded our method into a state-of-the-art text-spotting method and enhanced its robustness against image damage. 
We also built a series of baseline methods and conducted extensive discussions and comparisons, which demonstrated the advantages and effectiveness of our method.

\bibliographystyle{splncs04}
\bibliography{main}
\end{document}

%% file: gqwritepackage.tex
\usepackage{epsfig}
\usepackage{graphicx}
\usepackage{amsmath}
\usepackage{amssymb}

\usepackage{multirow}
\usepackage{tabularx}
\usepackage{booktabs}
\usepackage{longtable}
\usepackage{makecell}

\usepackage{dsfont}
\usepackage{pifont}
\usepackage{url}
\usepackage{color}
\usepackage[export]{adjustbox}
\usepackage{comment}
\usepackage{algorithm} 

\usepackage[utf8]{inputenc} 
\usepackage[T1]{fontenc}    
\usepackage{url}            
\usepackage{amsfonts}       
\usepackage{nicefrac}       
\usepackage{enumitem}
\usepackage{blindtext}
\usepackage{sidecap}
\usepackage{wrapfig}
\usepackage{dsfont}

\definecolor{dg}{rgb}{0,0.694,0.298}
\definecolor{purple}{rgb}{0.4,0.176,0.569}
\definecolor{royalblue}{RGB}{65,105,225}
\usepackage{pifont}
\newcommand{\figref}[1]{Fig.~\ref{#1}}
\newcommand{\reqref}[1]{Eq.~(\ref{#1})}
\newcommand{\secref}[1]{Sec.~\ref{#1}}
\newcommand{\tableref}[1]{Table~\ref{#1}}

\makeatletter
\DeclareRobustCommand\onedot{\futurelet\@let@token\@onedot}
\def\@onedot{\ifx\@let@token.\else.\null\fi\xspace}
\def\eg{\emph{e.g}\onedot} 
\def\ie{\emph{i.e}\onedot}

\def\etal{\emph{et al}\onedot}
\makeatother

\definecolor{americanrose}{rgb}{1.0, 0.01, 0.24}

